\documentclass{article}
\usepackage{arxiv}

\usepackage[utf8]{inputenc} 
\usepackage[T1]{fontenc}    
\usepackage{hyperref}       
\usepackage{url}            
\usepackage{booktabs}       
\usepackage{amsfonts}       
\usepackage{nicefrac}       
\usepackage{microtype}      
\usepackage{lipsum}
\usepackage{graphicx}
\usepackage{amsmath}
\usepackage{pifont}
\graphicspath{ {./images/} }

\title{Flow and Depth Assisted Video Prediction with Latent Transformer}

\author{
  Eliyas Suleyman \\
  School of Coumputing Science\\
  University of Glasglow\\
  Glasgow, Scotland, UK, G12 8QQ\\
  \texttt{2683522s@student.gla.ac.uk} \\
   \And
  Paul Henderson \\
  School of Coumputing Science\\
  University of Glasglow\\
  Glasgow, Scotland, UK, G12 8QQ\\
  \texttt{paul.henderson@glasgow.ac.uk} \\
  \And
  Eksan Firkat \\
  Tsinghua Shenzhen International Graduate School\\
  Tsinghua University\\
  Shenzhen, China\\
  \texttt{eksan@mail.tsinghua.edu.cn} \\
  \And
   Nicolas Pugeault \\
  School of Coumputing Science\\
  University of Glasglow\\
  Glasgow, Scotland, UK, G12 8QQ\\
  \texttt{nicolas.pugeault@glasgow.ac.uk} \\
}

\begin{document}

\maketitle

\begin{abstract}
Video prediction is a fundamental task for various downstream applications, including robotics and world modeling. 
Although general video prediction models have achieved remarkable performance in standard scenarios, occlusion is still an inherent challenge in video prediction.
We hypothesize that providing explicit information about motion (via point-flow) and geometric structure (via depth-maps) will enable video prediction models to perform better in situations with occlusion and the background motion.
To investigate this, we present the first systematic study dedicated to occluded video prediction. 
We use a standard multi-object latent transformer architecture to predict future frames, but modify this to incorporate information from depth and point-flow.
We evaluate this model in a controlled setting on both synthetic and real-world datasets with not only appearance-based metrics but also Wasserstein distances on object masks, which can effectively measure the motion distribution of the prediction.
We find that when the prediction model is assisted with point flow and depth, it performs better in occluded scenarios and predicts more accurate background motion compared to models without the help of these modalities.\footnote{This paper is accepted to British Machine Vision Conference (BMVC) 2025 Smart Camera Workshop}
\end{abstract}

\section{Introduction}

Video prediction is a crucial task for intelligent agents, with applications in robotics \cite{hu2024video}, autonomous driving \cite{Yang_2024_CVPR}, world models \cite{wang2024worlddreamer}, and weather forecasting \cite{nature-weather}. Accurately predicting the near future (e.g., 1–2 seconds) is essential, as it directly improves an agent’s decision-making efficiency~\cite{bharadhwaj2024track2act}.
The development of vision transformers for images \cite{dosovitskiy2020image} and videos \cite{arnab2021vivit} have made possible to improve video prediction quality \cite{ma2024latte, rakhimov2020latent, ye2023video, singh2024gsstu}.
However, due to the inherent complexity of motion in dynamic scenes with multiple objects, occlusions frequently occur, and latent transformer models can still struggle to accurately estimate the motion of objects that become temporarily invisible \cite{scat}.

Several approaches aim to improve video prediction by incorporating optical flow estimation \cite{bei2021learning, lu2021video, luo2021future, zhang2024extdm}. 
However, two major limitations of optical flow are that it accumulates errors over time, and loses information when objects become fully occluded.
As a result, optical flow-based methods struggle to handle complete occlusions effectively.
Unlike optical flow, which relies on dense pixel-wise motion estimation, recent progress in point tracking methods enable more robust occlusion handling by tracking and estimating key points on objects even when they are fully occluded \cite{karaev2025cotracker, tumanyan2024dino, xiao2024spatialtracker}. Furthermore, background motion is also well represented by point-flow, which is essential for modeling the motion of an ego-camera (e.g., autonomous cars).
Equally critical to occlusion handling, depth maps can provide geometric structure of the scene, allowing for better spatial reasoning in occlusion scenarios.

In this work, we hypothesize that integrating information about depth and the flow of points into a video prediction model will enhance its ability to anticipate object and background motion, particularly in occluded scenarios.
While point-flow helps track object motion trajectories, depth maps introduce explicit spatial constraints that improve occlusion-aware prediction.
To investigate this, we use latent transformer as our video prediction model, which lacks robustness to occlusions when only relying on RGB images \cite{scat}, and propose a variant that incorporates both point-flow and depth as additional modalities. 
Our approach enables the model to retain motion information when objects become temporarily invisible, improving future frame prediction accuracy by leveraging both motion trajectories and spatial structure alongside visual cues.
Furthermore, with the assistance of point-flow, the background motion can be predicted more accurately with more precise direction of background motion.

Our main contributions are as follows:  
\begin{itemize}
    \setlength{\itemsep}{-2pt}
    \item We provide the first systematic analysis of how depth and point-flow impact the performance of prediction when dynamic scenes have occlusions and background motion. 
    \item We design a video prediction model that can incorporate point-flow and depth as additional modalities to improve RGB frame prediction.
    \item We conduct extensive experiments on both synthetic and real world datasets, with several model variants and baselines.
    \item We find that when integrating point flow, the reappearance of occluded objects and the background motion are predicted more accurately.
\end{itemize}

\section{Related Work}


\paragraph{General Video Prediction Methods.}
Various neural network architectures have been explored for video prediction: hybrid models that combine RNNs and CNNs \cite{wang2022predrnn, gao2022simvp, chang2021mau}; latent transformer \cite{vaswani2017attention} models \cite{yan2021videogpt, wu2024ivideogpt, scat}, where the latent is usually encoded by a VQ-VAE or VQ-GAN \cite{van2017neural, razavi2019generating, esser2021taming} then the prediction is conducted by a transformer; diffusion \cite{ho2022video}, latent-diffusion \cite{rombach2022high} and diffusion-transformer \cite{peebles2023scalable} based approaches \cite{liu2024sora, yin2023nuwa}. 
Although the overall performance of these approaches is promising, since they lack explicit object or motion information, learning to generate complex motion is very expensive in terms of data and compute.

\paragraph{Optical Flow in Video Prediction.}
Optical flow is a pixel-wise dense motion estimation between consecutive video frames. 
FlowNet \cite{dosovitskiy2015flownet} and its advanced version \cite{ilg2017flownet} is first introduced to estimate the optical flow through CNN network. 
Recent optical flow estimation approaches used vision transformers to achieve the same goal \cite{shi2023flowformer++, le2024dense, lu2023transflow}. 
Because optical flow contains rich motion information of a dynamic scene, it is integrated to many video prediction approaches to predict future frames.
Li et al \cite{li2018flow} first predict the optical flow of future frames by condition on a single frame, then warp the RGB frame with predicted flow to achieve video prediction.
Shi et al \cite{shi2024motion} used a similar idea to predict the flow first then use a diffusion model conditioned on flow to generate RGB frames.
Bei et al \cite{bei2021learning} proposed a semantic aware approach that predicts the optical flow directly with a ConvLSTM network, then uses the predicted flow to generate the future frames.
Wu et al \cite{wu2022optimizing} used optical flow to optimize the model's frame interpolation ability to improve the future frame prediction quality.
Liang et al \cite{liang2024flowvid} generated video frames based on another video's optical flow information.
Optical flow has also been integrated with generative diffusion models to guide the motion of generated frames to be more realistic \cite{chefer2025videojam}.
However, error accumulation over time and the complete loss of information while objects are occluded hampers the effectiveness of optical flow methods when occlusion occurs.

\paragraph{Point Tracking.}
Point tracking approaches have recently gained popularity due to their strong performance \cite{karaev2025cotracker, tumanyan2024dino, cho2024flowtrack, xiao2024spatialtracker}.
Unlike optical flow estimation, which aims to estimate the motion of every pixel in an image, point tracking methods typically operate in an encoded latent space and focus on tracking sparse, semantically meaningful features.
Rather than modeling dense pixel-level motion, these methods estimate the trajectories of key features across frames, making them more robust to noise, occlusions, and appearance changes.
This abstraction allows tracking-based approaches to better capture high-level motion dynamics and structural consistency compared to traditional flow-based methods.
Several studies have attempted to integrate point tracking for motion modeling and future trajectory prediction.
For instance, \cite{bharadhwaj2024track2act} leveraged point tracking to assist robotic arm control in completing various tasks, achieving superior performance.
Point tracking has also been applied to generative tasks. 
\cite{jeong2024track4gen} incorporated point tracking into video diffusion models, enabling more realistic motion generation.


\section{Methodology}
\begin{figure*}
    
    \centering
    \includegraphics[trim=0 635 0 0, clip, width=\textwidth, keepaspectratio]{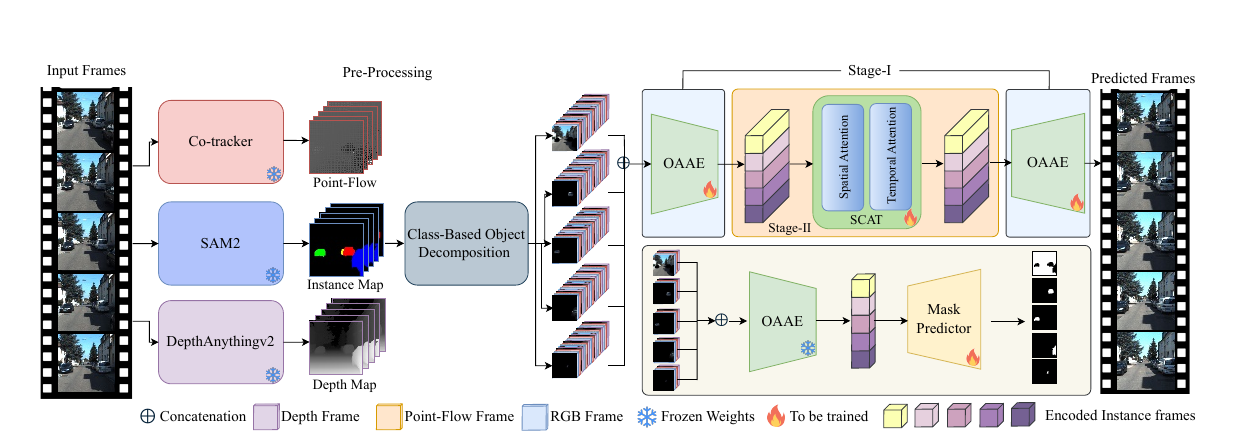} \label{fig:full-model}
    {\tiny\caption{\textbf{The overview of the proposed method.} First we obtain different modalities by using Cotracker and DepthAnythingV2; then we use SAM2 to segment the original RGB frames sequence to decompose the objects, segmentation map from SAM2 is also used to decompose the point-flow and depth map; After preprocessing, we first train OAAE to convert the frames into a latent space; then we train SCAT to predict the future latent frames; finally the predicted latent future frames are reconstructed by trained OAAE; The lower right box shows how we train a object mask predictor based on trained OAAE's latent space; after mask predictor is trained, it is then used solely for evaluating EMD.} 
    \vspace{-15pt}}
    \label{fig:overview}
\end{figure*}
\subsection{Preliminaries}


Let $X^{1:T}=\langle x^1,x^2,...,x^T \rangle$, be a sequence of $T$ RGB frames from a video clip, where $x^t\in\mathbb{R}^{h \times w \times 3}$. 
Our goal is to learn a probability distribution on future frames $X^{T+1:T+M}$, conditioned on the past frames $X^{1:T}$.
We next discuss the base model we build on in this paper, as well as the models used to extract additional modalities---point-flow and depth.

\paragraph{Base Architecture.}
We use the Stochastic Class-Attended Transformer (SCAT) \cite{scat} as the backbone structure for our model.
SCAT is a recent latent-transformer-based approach designed for object-centric video prediction. 
It is a two-staged approach that first trains an object aware auto-encoder (OAAE) to encode the video frames into latent representations. 
Then, a GPT-style transformer is trained on the past latent frames to predict future latent frames.
Finally, the predicted latent frames are decoded via OAAE to reconstruct the predicted frames.
SCAT offers a favorable trade-off between performance and computational cost, using a relatively lightweight transformer module for temporal prediction;
it achieves higher accuracy for similar parameter count versus similar non-object-centric models.

In the first stage, each frame $x$ from a video clip is decomposed by an off-the-shelf instance segmentation model \cite{ravi2024sam2} yielding $I$ number of instances with $m$ number of classes. 
Each instance is extracted by applying the corresponding mask predicted by the instance segmentation model. 
The full frame can be reconstructed by adding all of the masked instances, i.e.~%
 $x = \sum_{k=1}^{I} {\tilde{x}_{k}}$, 
where $\tilde{x}_{k}$ represents the $k^{th}$ instance.
Then, each instance is encoded by a set of class-specific encoders, denoted as \(\Phi = \{ \phi_1,\phi_2,...,\phi_m \}\), each of which encodes instances belonging to one class.
Then the encoded latent is quantized by class-specific embedding code books $E = \{ e_1,e_2,...,e_m\}$.
Each quantized instance is concatenated to generate a structured latent representation \( z \) that captures the full frame. 
A joint decoder $\Psi$ then reconstructs the original video frame \( x \) from the latent representation \( z \).
To enhance feature extraction and improve performance, we replace the encoder used in SCAT with SlotDiffusion's frame encoder \cite{wu2023slotdiffusion}, which uses a ResNet-18 \cite{resnet} architecture.
This modification provides stronger feature representations, enabling more effective encoding of object-centric information.
Moreover, the latent space is significantly smaller than SCAT's while maintaining similar or better performance

After the OAAE is trained, video clips are converted from RGB images into latent representation using OAAE.
Since each frame contains structured information of the instances, we can represent instances sequence as $\tilde{Z}_k=\{z_k^1, z_k^2,...,z_k^T | k=1, 2,...,I\}$, then the sequence representing the full scene can also obtained additively by summing all instance sequences as  
 $Z = \sum_{k=1}^{N} {\tilde{Z}_{k}}$, 
In second stage, SCAT uses class-specific transformer blocks for each semantic class, similarly to the first stage.
It models the motion pattern of an instance $k$ with self-attention individually (eq. \ref{eq:self_att}), as well as the potential relationship with other instances via cross-attention (eq. \ref{eq:self_att}) as shown in the equation blow:
\begin{equation}\label{eq:self_att}
\operatorname{SA}_{c}(\tilde{Z}_{k}) = \operatorname{softmax}\left(\frac{Q_{k}{K_k}^{T}}{\sqrt{d_{k}}}\right)V_k, \quad
\operatorname{CA}(\tilde{Z}_k) = \bigoplus_{i=1,\ldots,N,\, i \neq k} \operatorname{softmax}\left(\frac{Q_{k}{K_i}^{T}}{\sqrt{d_{k}}}\right)V_i
\end{equation}
where $\bigoplus$ denotes concatenation operation; 
The cross-attention layer's output, being \(I-1\) times larger than the input because of concatenation, is reduced to the original size through a linear layer.
Additionally, each transformer block's attention mechanism further operates on spatial and temporal dimensions to effectively capture both spatial and temporal dependencies, following \cite{bertasius2021space}, both within instance $k$ and across other instances. 
The final output is a probability distribution over OAAE codebook indices for each instance in each future frame.

\paragraph{Point Tracking with Cotracker.}
Cotracker \cite{karaev2025cotracker} is a transformer-based model that tracks 2D points in video sequences. 
First, the query points are initialized on the first frame of a video clip, with their initial positions and visibility.
A point $P_i$ at time step $t$ is represented as $ P_i^t =(x_i^t, y_i^t) \in \mathbb{R}^2, \text{for } t \in \{1, \ldots, T\}$.
It is set to make all points visible after it is initialized at the first time step (e.g first frame of a video clip) to reduce ambiguity.
After the points are initialized, an end-to-end CNN network is trained to obtain the feature map of the frames. 
Then each point is projected to the relative position on the feature map, and the corresponding feature is selected for the point.
Finally, a transformer model is trained iteratively to learn how these points and selected feature maps are correlated.
The objective of this model is to minimize the distance between the predicted and ground truth point locations.

\paragraph{Depth Estimation with DepthAnything-V2.}
Depth Anything \cite{yang2024depth,yang2024depthv2} is a monocular depth estimation model designed to generalize well across diverse real-world scenes.
It follows a semi-supervised learning approach, where a teacher-student framework is employed to leverage both synthetic and real data.
Initially, a teacher network is trained on a large-scale synthetic dataset with dense ground-truth depth annotations. 
This teacher is then used to pseudo-label a large corpus of real-world unlabeled images, effectively transferring its knowledge to real data. 
Finally, a student network is trained on a mixture of these pseudo-labeled real images and a small set of manually labeled real-world samples.
The model takes a single RGB frame as input and produces a dense depth map as output.
We use the second version as our depth estimator for video frames.

\subsection{Proposed Method}
SCAT \cite{scat} decomposes a video into object instances, and an instance that becomes completely occluded at a certain time step is not visible to the encoder.
This makes it difficult to predict the motion of fully occluded objects, even when explicit visual information about these objects is available.
We therefore propose incorporating tracked points from Cotracker as point-flows, providing additional information to the prediction model.
Point-flows offer a more effective and robust alternative to optical flow for achieving object tracking, as optical flow tends to accumulate errors over time \cite{harley2022particle}. 
By utilizing point-flows, the encoder can retain information about an instance's position at a certain time step $t$, even when its RGB image is entirely absent due to complete occlusion.

We hypothesize that incorporating point-flows alongside RGB frames during encoding will enrich the latent representations with relative location information.
Therefore, the motion of occluded objects can be predicted more accurately.
Depth images are integrated as a another modality to our model, providing geometric context that is invariant to appearance changes.
While point flows capture motion, depth encodes scene structure, aiding in disambiguating object movement and handling occlusions—especially under camera motion—thus improving spatial and temporal reasoning.
It is important to note that we do not require any additional or richer information to train our model. 
Instead, we use pretrained models solely to preprocess the available RGB sequences, generating point-flow and depth images from the same input data used by existing baselines.
Following SCAT, we test our hypothesis by designing a family of models with varying input configurations:
\textbf{SCAT-D}, trained with RGB frames and depth frames;
\textbf{SCAT-P}, trained with RGB frames and point-flows; 
and 
\textbf{SCAT-DP} trained with RGB frames, depth images, and point-flows. 

\paragraph{Point flow and Depth.}
We first use Cotracker to track points in a video clip, then calculate the point-flow as the displacements of each point between consecutive frames.
For the initial time step (\(t = 0\)), there are no displacements, as the points are treated as the initial reference positions, represented by a tensor of shape \((T, N, 3)\), where \(T\) is the number of frames, \(N\) is the number of points, and \(3\) represents the \((h, w)\) coordinates and visibility.
From the second frame and onward (\(t \geq 1\)), the horizontal and vertical displacements of each point are calculated as the difference between the current and previous positions. 
Finally, since each point is defined by its \((h, w)\) coordinates, the displacement information is mapped to a grid with the same size as the image, resulting in a tensor of shape \((T, H, W, 3)\), where \(H\) and \(W\) represent the height and width of the video frame resolution. 
The last dimension encodes horizontal displacement, vertical displacement, and visibility.
We therefore have
\begin{equation}
\begin{split}
\mathbf{PointFlow}(T, H, W, 3) =
\begin{cases} 
(0, 0, 1), & \text{if } t = 0, \\
(h_{t}^{n} - h_{t-1}^{n}, w_{t}^{n} - w_{t-1}^{n}, v_{t}^{n}), & \text{if } t > 0.
\end{cases}
\end{split}
\end{equation}
where $\mathbf{PointFlow}(T, H, W, 3)$ is the displacement tensor, \(x_{t, n}\) and \(y_{t, n}\) are the \((h, w)\) coordinates of the \(n^{th}\) point and \(v_{t, n}\) is the visibility of the \(n^{th}\) point at time step \(t\).
\((H, W)\) corresponds to the pixel grid location in the image, derived from the \((h, w)\) coordinates of each point. 
This mapping ensures that the point-flows retain spatial correspondence with the video frames, enabling effective integration with the encoder.

For depth images, we employ an off-the-shelf depth estimation model \cite{yang2024depth} to generate the depth information for non-synthesized datasets.
Since a video sequence is composed of instance sequences, the corresponding points and depth information are extracted via segmentation map used to decomposed the instances. 

After we obtain these modalities, we concatenate them with the original RGB frame on the channel dimension to form the input of the encoder.
Then, all of these information will be encoded together according to different variants of our proposed method.
Finally, the model's output is not just a single RGB frames but as well as other modalities.
This makes sure that other modalities will be encoded to the latent space.

\paragraph{Loss Function.}
Since our approach has two stages, we need to train the frame encoder first and then train the temporal predictor.
For the frame encoder, we modify the original VQ-loss and Commitment Loss to fit our model design.
We extend VQ loss for each semantic class separately because each instance is encoded via a class-specific encoder and codebook, then the overall reconstruction loss for RGB images, depths and point-flows is calculated. $L_{VQ}, L_{recon}$ is shown below:
\[
\mathcal{L}_{VQ} = \sum_{c=1}^{m}\sum_{k=1}^{n_c} \left\Vert \operatorname{sg}[\tilde{z}^{c}_{k}] - e_c \right\Vert_{2}^{2}, \quad
\mathcal{L}_{commitment} = \sum_{c=1}^{m}\sum_{k=1}^{n_c} \left\Vert \tilde{z}^{c}_{k} - \operatorname{sg}[e_c] \right\Vert_{2}^{2}
\]
\begin{equation}
\mathcal{L}_{recon} = -\log{p(x|\Psi(\Phi(x)))}
\end{equation}
where $\operatorname{sg}$ denotes the stop-gradient operator, $n_c$ represents the number of instances in class $c$, and $e_c$ corresponds to the codebook for class $c$, respectively.
We also include LPIPS \cite{lpips} as an additional reconstruction loss:
\begin{equation}
    \mathcal{L}_{\text{LPIPS}}(x, \Psi(\Phi(x))) = \sum_l w_l \left\| \phi_l(x) - \phi_l(\Psi(\Phi(x))) \right\|_2^2
\end{equation}
where $\phi_l(x)$ represents the deep feature maps extracted from the $l$-th layer of a pretrained network $\phi$. The term $w_l$ is a learned weight that adjusts the contribution of each layer to the overall similarity, and $\|\cdot\|_2^2$ denotes the squared Euclidean distance between feature representations.
The final objective of our encoder will be summing all loss terms together as $\mathcal{L}=\mathcal{L}_{VQ}+\mathcal{L}_{commitment}+\mathcal{L}_{recon}+\mathcal{L}_{\text{LPIPS}}$.

For the transformer model that predicts future frames in latent space, we use the same formulation as SCAT, i.e.~minimizing the cross entropy between target and predicted indices.

\section{Experiments} \label{experiments}

We first conduct a series of experiments to analyze the impact of each additional modality on future frame prediction using the proposed family of models.
Our primary focus is on evaluating occluded scenarios under controlled settings, enabling a systematic assessment of how well each modality improves performance in handling occlusions.
We focus our evaluation on the predicted RGB frames and moving object's mask but not the other modalities which are simply regarded as guidance for the model.
To demonstrate the generality of the proposed method, we also evaluate it on more diverse scenarios and compare its performance against other baselines.
In each experiment, we follow SCAT's experimental setups, where the proposed model is required to predict 5 to 20 future frames given five input frames. 
All experiments are conducted on a single NVIDIA RTX 3090 GPU, and the model sizes (e.g., number of parameters) of other baselines are adjusted accordingly to ensure a fair comparison.

\subsection{Datasets}
\paragraph{Kubric Occlusion:} The hypothesis of this paper is that incorporating point-flow can improve the performance of prediction models, particularly in scenarios involving occlusions. 
To test this, we used Kubric \cite{greff2022kubric} to generate video clips tailored for our evaluation, which we refer as \textbf{Kubric-Occlusion}.
A total of 1,800 video clips were generated, with 1,300 used for training and 500 for testing. In each clip, one object remains stationary at a random location, while another object appears at a random position and moves behind the stationary object, creating an occlusion event.

\paragraph{KITTI:} The \textbf{KITTI dataset} \cite{geiger2013vision} is a widely used benchmark for autonomous driving research. It contains diverse driving scenarios captured in urban, residential, and highway environments. 
In this work, we use a \textbf{subset} of KITTI, specifically selecting scenes from \textit{city, residential, and road} categories. 
We first preprocess the dataset to obtain all of the car instances;
then, we sort the segmented car instances by size and select the largest four as foreground objects; the remainder of the image is categorized as background. 
After processing, 2,497 clips are used as training and 639 for testing (each clip contains 10 frames).

\subsection{Evaluation Metrics}
We evaluate the pixel-level quality of predicted frames using standard metrics: PSNR\cite{PSNR}, LPIPS\cite{zhang2018unreasonable}, and SSIM\cite{SSIM}. 
However, since the primary focus of our work is on assessing motion in the predicted frames, appearance-based metrics alone are insufficient to capture the dynamic aspects of prediction quality. 
To address this, we introduce the optical flow difference (OFD), which measures the discrepancy in motion between predicted and ground truth frames.
Optical flow is computed using the Gunnar-Farneback method~\cite{farneback2003two}, the motion accuracy is then quantified by calculating the mean squared error ($L_2$ loss) between the predicted and ground truth flows.

In addition to global motion assessment via OFD, we further evaluate motion quality at the instance level. 
We train a mask predictor to predict instance masks from the trained VQ-VAE latent space, and use this to estimate masks for predicted frames.
We then compute the Earth Mover’s Distance (EMD) (also known as Wasserstein distance) between the ground truth masks.
While OFD captures overall scene motion, EMD provides a finer-grained analysis of motion distribution differences, offering a more accurate reflection of motion quality in predicted frames.
EMD we use in our paper is defined as follows:
Let \( P_t = \{ \mathbf{p}_1, \dots, \mathbf{p}_m \} \subset \mathbb{R}^2 \) be the set of pixel coordinates for the predicted mask,
and \( G_t = \{ \mathbf{g}_1, \dots, \mathbf{g}_n \} \subset \mathbb{R}^2 \) be the set of pixel coordinates for the ground truth mask.
We define uniform discrete distributions over these sets:
$
\mathbf{a} = \left( \frac{1}{m}, \dots, \frac{1}{m} \right) \in \Delta^m, \quad
\mathbf{b} = \left( \frac{1}{n}, \dots, \frac{1}{n} \right) \in \Delta^n
$
Let \( M \in \mathbb{R}^{m \times n} \) be the cost matrix with entries: $ M_{ij} = \| \mathbf{p}_i - \mathbf{g}_j \|_2 $
The Earth Mover’s Distance (squared Wasserstein distance) is computed as the optimal transport cost:
\[
\text{EMD}^2(P, G) = \min_{T \in U(\mathbf{a}, \mathbf{b})} \sum_{i=1}^{m} \sum_{j=1}^{n} T_{ij} M_{ij}
\]
where \( U(\mathbf{a}, \mathbf{b}) = \{ T \in \mathbb{R}_+^{m \times n} \mid T \mathbf{1}_n = \mathbf{a},\; T^\top \mathbf{1}_m = \mathbf{b} \} \) is the set of admissible transport plans.
All metrics are computed on a per-frame basis, and the values reported in the table represent the mean over all frames across the clips in the respective dataset.

\subsection{Results}
In this section, we first evaluate the proposed family of models by comparing their performance internally on occluded and general scenarios with backround motion.
Then we select the best performing model to compare against other similar approaches.
The goal is to analyze the contribution of each additional modality (point-flow, depth, or both) and determine which variant performs best under different conditions. 
By conducting these internal experiments, we aim to identify the most effective model configuration before benchmarking it against other similar approaches.
\begin{figure}
  \centering
  \small 
  \setlength{\tabcolsep}{0.5pt} 
  \begin{tabular}{c cc|cccccc} 
    & \multicolumn{2}{c}{\centering{Input}} &   \multicolumn{6}{c}{\centering{Prediction}}   \\
    & \textbf{$t=1$} & \textbf{$t=5$} & \textbf{$t=13$} & \textbf{$t=14$} & \textbf{$t=15$} & \textbf{$t=16$} & \textbf{$t=17$}\\
    \rotatebox{90}{GT} &            \includegraphics[width=0.065\textwidth]{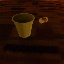} &
                                    \includegraphics[width=0.065\textwidth]{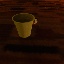} &
                                    \includegraphics[width=0.065\textwidth]{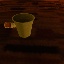} &
                                    \includegraphics[width=0.065\textwidth]{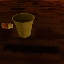} &
                                    \includegraphics[width=0.065\textwidth]{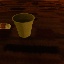} &
                                    \includegraphics[width=0.065\textwidth]{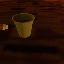} &
                                    \includegraphics[width=0.065\textwidth]{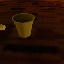} & \\
     & \multicolumn{2}{c}{SCAT}    &
                                    \includegraphics[width=0.065\textwidth]{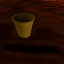} &
                                    \includegraphics[width=0.065\textwidth]{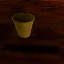} &
                                    \includegraphics[width=0.065\textwidth]{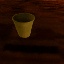} &
                                    \includegraphics[width=0.065\textwidth]{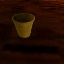} &
                                    \includegraphics[width=0.065\textwidth]{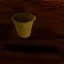} & \\
     & \multicolumn{2}{c}{SCAT-P}    &
                                    \includegraphics[width=0.065\textwidth]{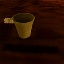} &
                                    \includegraphics[width=0.065\textwidth]{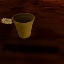} &
                                    \includegraphics[width=0.065\textwidth]{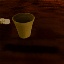} &
                                    \includegraphics[width=0.065\textwidth]{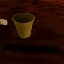} &
                                    \includegraphics[width=0.065\textwidth]{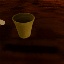} & \\
    & \multicolumn{2}{c}{SCAT-D}   &
                                    \includegraphics[width=0.065\textwidth]{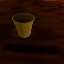} &
                                    \includegraphics[width=0.065\textwidth]{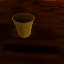} &
                                    \includegraphics[width=0.065\textwidth]{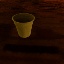} &
                                    \includegraphics[width=0.065\textwidth]{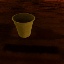} &
                                    \includegraphics[width=0.065\textwidth]{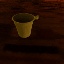} & \\
    & \multicolumn{2}{c}{SimVP}   &
                                    \includegraphics[width=0.065\textwidth]{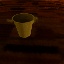} &
                                    \includegraphics[width=0.065\textwidth]{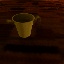} &
                                    \includegraphics[width=0.065\textwidth]{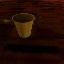} &
                                    \includegraphics[width=0.065\textwidth]{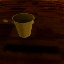} &
                                    \includegraphics[width=0.065\textwidth]{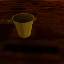} & \\
    & \multicolumn{2}{c}{SCAT-PD}   &
                                    \includegraphics[width=0.065\textwidth]{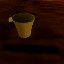} &
                                    \includegraphics[width=0.065\textwidth]{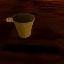} &
                                    \includegraphics[width=0.065\textwidth]{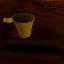} &
                                    \includegraphics[width=0.065\textwidth]{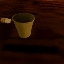} &
                                    \includegraphics[width=0.065\textwidth]{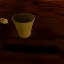} & \\
  \end{tabular}
  \begin{tabular}{c cc|cccccc} 
    & \multicolumn{2}{c}{\centering{Input}} &   \multicolumn{6}{c}{\centering{Prediction}}   \\
    & \textbf{$t=1$} & \textbf{$t=5$} & \textbf{$t=6$} & \textbf{$t=7$} & \textbf{$t=8$} & \textbf{$t=9$} & \textbf{$t=10$}\\
    \rotatebox{90}{GT} &            \includegraphics[width=0.065\textwidth]{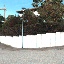} &
                                    \includegraphics[width=0.065\textwidth]{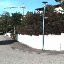} &
                                    \includegraphics[width=0.065\textwidth]{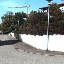} &
                                    \includegraphics[width=0.065\textwidth]{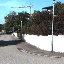} &
                                    \includegraphics[width=0.065\textwidth]{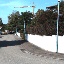} &
                                    \includegraphics[width=0.065\textwidth]{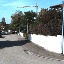} &
                                    \includegraphics[width=0.065\textwidth]{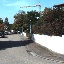} & \\
     & \multicolumn{2}{c}{SCAT}    &
                                    \includegraphics[width=0.065\textwidth]{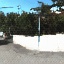} &
                                    \includegraphics[width=0.065\textwidth]{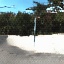} &
                                    \includegraphics[width=0.065\textwidth]{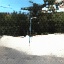} &
                                    \includegraphics[width=0.065\textwidth]{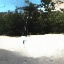} &
                                    \includegraphics[width=0.065\textwidth]{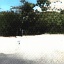} & \\
     & \multicolumn{2}{c}{SCAT-P}    &
                                    \includegraphics[width=0.065\textwidth]{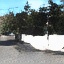} &
                                    \includegraphics[width=0.065\textwidth]{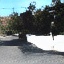} &
                                    \includegraphics[width=0.065\textwidth]{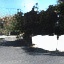} &
                                    \includegraphics[width=0.065\textwidth]{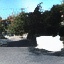} &
                                    \includegraphics[width=0.065\textwidth]{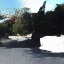} & \\
    & \multicolumn{2}{c}{SCAT-D}   &
                                    \includegraphics[width=0.065\textwidth]{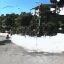} &
                                    \includegraphics[width=0.065\textwidth]{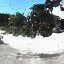} &
                                    \includegraphics[width=0.065\textwidth]{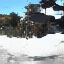} &
                                    \includegraphics[width=0.065\textwidth]{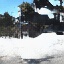} &
                                    \includegraphics[width=0.065\textwidth]{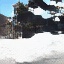} & \\
    & \multicolumn{2}{c}{SimVP}   &
                                    \includegraphics[width=0.065\textwidth]{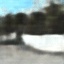} &
                                    \includegraphics[width=0.065\textwidth]{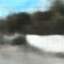} &
                                    \includegraphics[width=0.065\textwidth]{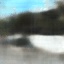} &
                                    \includegraphics[width=0.065\textwidth]{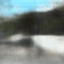} &
                                    \includegraphics[width=0.065\textwidth]{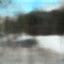} & \\
    & \multicolumn{2}{c}{SCAT-PD}   &
                                    \includegraphics[width=0.065\textwidth]{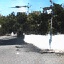} &
                                    \includegraphics[width=0.065\textwidth]{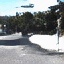} &
                                    \includegraphics[width=0.065\textwidth]{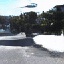} &
                                    \includegraphics[width=0.065\textwidth]{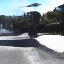} &
                                    \includegraphics[width=0.065\textwidth]{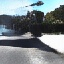} & \\
  \end{tabular}
  \caption{Comparison of different model variants on the \textbf{Kubric-Occlusion (Left)} and \textbf{KITTI (Right)} dataset.
  }
  \label{fig:result-kubric}
  \vspace{-10pt}
\end{figure}
\begin{table*}
\centering \footnotesize
\setlength{\tabcolsep}{2pt}
\resizebox{\columnwidth}{!}{
\begin{tabular}{@{}lcccccc|cccccc@{}}
\toprule
        & \multicolumn{6}{c}{\textbf{Kubric-Occlusion}} &  \multicolumn{6}{c}{\textbf{KITTI}} \\ 
         & PSNR$\uparrow$ & SSIM$\uparrow$ & LPIPS$\downarrow$ & OFD$\downarrow$ &  EMD$\downarrow$ & Prms  & PSNR$\uparrow$ & SSIM$\uparrow$ & LPIPS$\downarrow$ & OFD$\downarrow$ &  EMD$\downarrow$ & Prms\\ \midrule
SCAT     & 25.88$\pm$0.13          & 0.658$\pm$0.007          & 0.064$\pm$0.001          & 0.0423$\pm$0.0017          & 0.0081$\pm$0.0005          & 11M 
         & 15.33$\pm$0.13          & \textbf{0.473$\pm$0.006} & 0.135$\pm$0.003          & 2.5776$\pm$0.3341          & 0.0310$\pm$0.0016          & 8M        \\ 
SCAT-P   & 25.99$\pm$0.13          & 0.665$\pm$0.007          & 0.063$\pm$0.001          & 0.0356$\pm$0.0016          & 0.0070$\pm$0.0006          & 11M 
         & 15.20$\pm$0.11          & 0.448$\pm$0.006          & 0.155$\pm$0.003          & 1.6659$\pm$0.1939          & 0.0282$\pm$0.0020          & 8M        \\ 
SCAT-D   & \textbf{26.53$\pm$0.13} & \textbf{0.701$\pm$0.006} & \textbf{0.054$\pm$0.001} & 0.0414$\pm$0.0019          & 0.0069$\pm$0.0004          & 11M 
         & \textbf{15.53$\pm$0.12} & 0.465$\pm$0.006          & \textbf{0.132$\pm$0.003} & 3.2781$\pm$0.4762          & 0.0285$\pm$0.0022          & 8M        \\ 
SCAT-PD  & 25.69$\pm$0.12          & 0.649$\pm$0.007          & 0.072$\pm$0.002          & \textbf{0.0347$\pm$0.0014} & \textbf{0.0066$\pm$0.0004} & 11M 
         & 15.36$\pm$0.11          & 0.445$\pm$0.006          & 0.137$\pm$0.002          & \textbf{1.6390$\pm$0.2324} & \textbf{0.0278$\pm$0.0016} & 8M        \\
\bottomrule
\end{tabular}
}              
\caption{Quantitative results on \textbf{Kubric-Occlusion} and \textbf{KITTI} dataset}
\label{tb:quantitative-kitti}
\label{tb:quantitative-kubric}
\end{table*}

From Table~\ref{tb:quantitative-kubric}, we can see that on the Kubric-occlusion dataset, all proposed variants improve on plain SCAT in terms of motion metrics.
This confirms our hypothesis that flow and depth modalities are important for occlusion prediction. Interestingly, the SCAT-D variant achieves the best performance for appearance metrics (PSNR, SSIM \& LPIPS), and SCAT-P achieves the best results for motion-relevant metrics (OFD \& EMD). We found the performance of the SCAT-PD variant to be generally lower than the two other (SCAT-P and SCAT-D), which is likely a consequence of processing larger input data with the same model size.
In Figure~\ref{fig:result-kubric}, we can also see the qualitative results reflect the quantitative scores:
The occluded object's reappearance is only predicted correctly when point flow is integrated (SCAT-P and -PD), confirming the evidence provided by the OFD and EMD metrics.  

%
\begin{table*}
\centering \small
\setlength{\tabcolsep}{2pt}
\resizebox{\columnwidth}{!}{
\begin{tabular}{@{}lcccccccccc@{}}
\toprule
          &               \multicolumn{5}{c}{\textbf{Kubric-Occlusion}} & \multicolumn{5}{c}{\textbf{KITTI}} \\
\cmidrule(r){2-6}  \cmidrule(r){7-11} 
          & PSNR$\uparrow$          & SSIM$\uparrow$           & LPIPS$\downarrow$        &  OFD$\downarrow$  &   Num-Params  &  PSNR$\uparrow$          & SSIM$\uparrow$          & LPIPS$\downarrow$        & OFD$\downarrow$    & Num-Params \\ \midrule
SCAT      & 25.88$\pm$0.13          & 0.66$\pm$0.007           & 0.064$\pm$0.001          & 0.0423$\pm$0.0017         &  11M  & 15.33$\pm$0.13           & 0.47$\pm$0.006          & \textbf{0.135$\pm$0.002} & 2.49$\pm$0.33           & 8M  \\ 
SimVP     & \textbf{33.05$\pm$0.13} & \textbf{0.95$\pm$0.001}  & \textbf{0.021$\pm$0.001} & 0.0626$\pm$0.0019         &  14M  &\textbf{17.14$\pm$0.10}   & \textbf{0.49$\pm$0.005} & 0.332$\pm$0.004          & 1.66$\pm$0.11           & 14M \\ 
Ours      & 25.69$\pm$0.12          & 0.65$\pm$0.007           & 0.072$\pm$0.002          & \textbf{0.0347$\pm$0.0014}&  11M  & 15.36$\pm$0.11           & 0.45$\pm$0.006          & 0.137$\pm$0.002          & \textbf{1.64$\pm$0.23}  & 8M  \\ \bottomrule
\end{tabular}}     
\caption{Comparison to previous works on \textbf{Kubric-Occlusion} and \textbf{KITTI} dataset}
\label{tb:quantitative-others}  
\end{table*}
Table \ref{tb:quantitative-others} provide a comparison to SimVP and plain SCAT. The proposed model appear to underperform SimVP when looking at appearance-based metrics on the \textbf{Kubric-Occlusion} dataset, however they perform much better when looking at motion-based metrics. This contrast can be explained by the comparatively small impact of moving objects on appearance metrics versus background noise, which is likely reduced by the larger size of the SimVP model.
This intuition is confirmed by the qualitative results shown in Figure \ref{fig:result-kubric}, where SCAT-P \& SCAT-DP predict accurately the motion of moving objects while others fail. Specifically, the trajectory of the moving object in Kubric-Occlusion (left) dataset is correctly predicted only when including point-flow information (SCAT-P \& SCAT-PD), while SimVP fails to predict the object's reappearance. 

In contrast, where KITTI features complex real world dynamics, our model outperforms SimVP in LPIPS (0.137 v 0.332).
Also, we see that in terms of motion our model also outperformed SimVP (1.64 v 1.66), this can be seen in Figure \ref{fig:result-kubric} (right).
The Figure \ref{fig:result-kubric} shows clear evidence that when the point-flow is integrated, the backgorund motion is predicted accurately (SCAT-P, D \& PD) versus RGB-only variants.  
It is important to note that our SCAT variants are nearly two times smaller than SimVP and achieved similar or better performance in terms of motion accuracy.



\section{Conclusion}
We propose a video prediction pipeline that investigates the impact of adding point tracking and depth information on future frame prediction.
Our method incorporates point-flow and depth maps to enhance motion prediction, particularly in challenging scenarios with occlusions and background motion.
Experimental results show that point-flow contributes to more accurate motion estimation, and in particular can successfully predict the reappearance of occluded moving objects. 
However, while adding multiple modalities improves general motion prediction, the additional input information can degrade pixel-level appearance quality when keeping model size constant.
In future work, we aim to explore strategies for a better integration of diverse modalities and improving reconstruction fidelity.
\bibliographystyle{unsrt}  
\bibliography{egbib}

\clearpage

\appendix

\section*{Autoencoder Performance Analysis}
We evaluate the performance of our Autoencoder in reconstructing RGB frames. 
As shown in Section \ref{experiments}, integrating only depth yields the best results in appearance-based metrics (PSNR, SSIM, LPIPS), but not in motion-based metrics (OFD, EMD).

To explore this further, we compare four variants of our Autoencoder by computing each metric based on the difference between reconstructed and ground-truth RGB frames.
\begin{table}[ht]
\centering \small
\setlength{\tabcolsep}{2pt}
\begin{tabular}{@{}lccccc@{}}
\toprule
& Depth & Point-Flow & PSNR$\uparrow$ & SSIM$\uparrow$ & LPIPS$\downarrow$ \\ \midrule
SCAT     & \ding{55} & \ding{55} & 26.672$\pm$0.133 & 0.669$\pm$0.007 & 0.052$\pm$0.001 \\
SCAT-P   & \ding{55} & \ding{51} & 26.737$\pm$0.129 & 0.677$\pm$0.006 & 0.052$\pm$0.001 \\
SCAT-D   & \ding{51} & \ding{55} & \textbf{27.280$\pm$0.127} & \textbf{0.712$\pm$0.006} & \textbf{0.043$\pm$0.001} \\
SCAT-PD  & \ding{51} & \ding{51} & 26.194$\pm$0.122 & 0.660$\pm$0.007 & 0.061$\pm$0.001 \\ \bottomrule
\end{tabular}
\caption{Autoencoder's frame reconstruction performance on \textbf{Kubric-Occlusion} dataset}
\label{tb:autoencoder-recon-kubric}
\end{table}
\begin{table}[ht]
\centering \small
\setlength{\tabcolsep}{2pt}
\begin{tabular}{@{}lccccc@{}}
\toprule
& Depth & Point-Flow & PSNR$\uparrow$ & SSIM$\uparrow$ & LPIPS$\downarrow$ \\ \midrule
SCAT     & \ding{55} & \ding{55} & \textbf{21.468$\pm$0.100} & 0.769$\pm$0.003          & \textbf{0.038$\pm$0.001} \\
SCAT-P   & \ding{55} & \ding{51} & 20.138$\pm$0.093          & 0.701$\pm$0.003          & 0.056$\pm$0.001          \\
SCAT-D   & \ding{51} & \ding{55} & 21.316$\pm$0.095          & \textbf{0.770$\pm$0.003} & 0.040$\pm$0.001          \\
SCAT-PD  & \ding{51} & \ding{51} & 19.957$\pm$0.090          & 0.695$\pm$0.003          & 0.063$\pm$0.001          \\ \bottomrule
\end{tabular}
\caption{Autoencoder's frame reconstruction performance on \textbf{KITTI} dataset}
\label{tb:autoencoder-recon-kitti}
\end{table}

In Kubric-Occlusion, we observe that integrating a single modality (SCAT-P or SCAT-D) improves overall reconstruction quality.
However, combining both modalities simultaneously (SCAT-PD) leads to decreased reconstruction performance. 
This suggests a trade-off between incorporating multiple modalities and reconstruction quality under a limited latent capacity.

In KITTI, Autoencoder performance significantly degrades when incorporating point flow (SCAT-P \& DP).
Unlike Kubric-Occlusion, KITTI involves camera motion, making the background non-stationary.
As a result, the loss function—tasked with reconstructing both RGB frames and additional modalities—introduces noise into the RGB output.
This is evident in the reconstructed frames, where small artifacts appear at point displacement regions.
These findings suggest that directly concatenating point-flow with RGB frames is not an effective encoding strategy.

\section*{More Qualitative Results}

\begin{figure}[th!]
  \centering
  \small 
  \setlength{\tabcolsep}{0.5pt} 
  \begin{tabular}{c|cccc} 
     GT &  SCAT & SCAT-P & SCAT-D & SCAT-DP   \\
     \includegraphics[width=0.17\textwidth]               {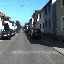} &
     \includegraphics[width=0.17\textwidth]           {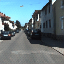} &
     \includegraphics[width=0.17\textwidth]      {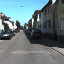} &
     \includegraphics[width=0.17\textwidth]     {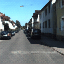} &
     \includegraphics[width=0.17\textwidth]{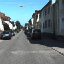} \\
     \includegraphics[width=0.17\textwidth]            {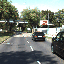} &
     \includegraphics[width=0.17\textwidth]           {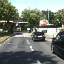} &
     \includegraphics[width=0.17\textwidth]      {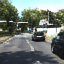} &
     \includegraphics[width=0.17\textwidth]     {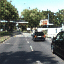} &
     \includegraphics[width=0.17\textwidth]{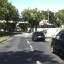} \\
     \includegraphics[width=0.17\textwidth]            {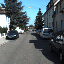} &
     \includegraphics[width=0.17\textwidth]           {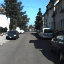} &
     \includegraphics[width=0.17\textwidth]      {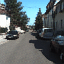} &
     \includegraphics[width=0.17\textwidth]     {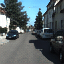} &
     \includegraphics[width=0.17\textwidth]{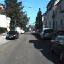} \\
     \includegraphics[width=0.17\textwidth]            {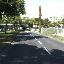} &
     \includegraphics[width=0.17\textwidth]           {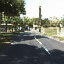} &
     \includegraphics[width=0.17\textwidth]      {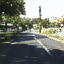} &
     \includegraphics[width=0.17\textwidth]     {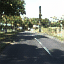} &
     \includegraphics[width=0.17\textwidth]{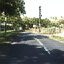} \\
     \includegraphics[width=0.17\textwidth]            {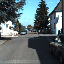} &
     \includegraphics[width=0.17\textwidth]           {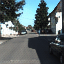} &
     \includegraphics[width=0.17\textwidth]      {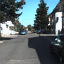} &
     \includegraphics[width=0.17\textwidth]     {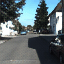} &
     \includegraphics[width=0.17\textwidth]{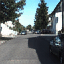} \\
     \includegraphics[width=0.17\textwidth]            {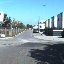} &
     \includegraphics[width=0.17\textwidth]           {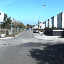} &
     \includegraphics[width=0.17\textwidth]      {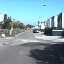} &
     \includegraphics[width=0.17\textwidth]     {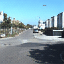} &
     \includegraphics[width=0.17\textwidth]{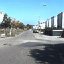} \\
     \includegraphics[width=0.17\textwidth]            {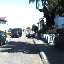} &
     \includegraphics[width=0.17\textwidth]           {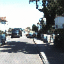} &
     \includegraphics[width=0.17\textwidth]      {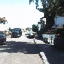} &
     \includegraphics[width=0.17\textwidth]     {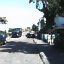} &
     \includegraphics[width=0.17\textwidth]{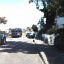} \\
     \includegraphics[width=0.17\textwidth]            {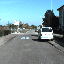} &
     \includegraphics[width=0.17\textwidth]           {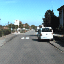} &
     \includegraphics[width=0.17\textwidth]      {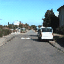} &
     \includegraphics[width=0.17\textwidth]     {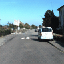} &
     \includegraphics[width=0.17\textwidth]{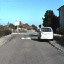} \\
  \end{tabular}
  \caption{Qualitative results on Autoencoder's reconstruction on KITTI dataset}
  \label{fig:result-kubric}
  \vspace{-10pt}
\end{figure}
\begin{figure}[th!]
  \centering
  \small 
  \setlength{\tabcolsep}{0.5pt} 
  \begin{tabular}{c|cccc} 
     GT &  SCAT & SCAT-P & SCAT-D & SCAT-DP   \\
     \includegraphics[width=0.17\textwidth]               {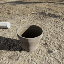} &
     \includegraphics[width=0.17\textwidth]           {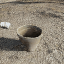} &
     \includegraphics[width=0.17\textwidth]      {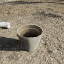} &
     \includegraphics[width=0.17\textwidth]     {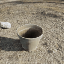} &
     \includegraphics[width=0.17\textwidth]{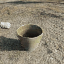} \\
     \includegraphics[width=0.17\textwidth]            {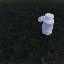} &
     \includegraphics[width=0.17\textwidth]           {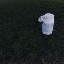} &
     \includegraphics[width=0.17\textwidth]      {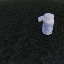} &
     \includegraphics[width=0.17\textwidth]     {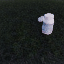} &
     \includegraphics[width=0.17\textwidth]{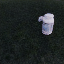} \\
     \includegraphics[width=0.17\textwidth]            {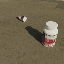} &
     \includegraphics[width=0.17\textwidth]           {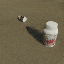} &
     \includegraphics[width=0.17\textwidth]      {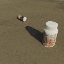} &
     \includegraphics[width=0.17\textwidth]     {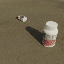} &
     \includegraphics[width=0.17\textwidth]{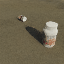} \\
     \includegraphics[width=0.17\textwidth]            {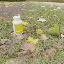} &
     \includegraphics[width=0.17\textwidth]           {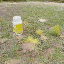} &
     \includegraphics[width=0.17\textwidth]      {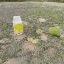} &
     \includegraphics[width=0.17\textwidth]     {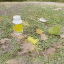} &
     \includegraphics[width=0.17\textwidth]{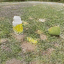} \\
     \includegraphics[width=0.17\textwidth]            {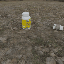} &
     \includegraphics[width=0.17\textwidth]           {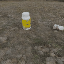} &
     \includegraphics[width=0.17\textwidth]      {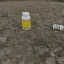} &
     \includegraphics[width=0.17\textwidth]     {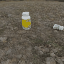} &
     \includegraphics[width=0.17\textwidth]{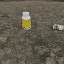} \\
     \includegraphics[width=0.17\textwidth]            {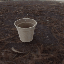} &
     \includegraphics[width=0.17\textwidth]           {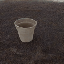} &
     \includegraphics[width=0.17\textwidth]      {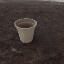} &
     \includegraphics[width=0.17\textwidth]     {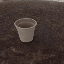} &
     \includegraphics[width=0.17\textwidth]{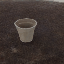} \\
     \includegraphics[width=0.17\textwidth]            {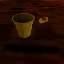} &
     \includegraphics[width=0.17\textwidth]           {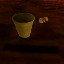} &
     \includegraphics[width=0.17\textwidth]      {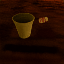} &
     \includegraphics[width=0.17\textwidth]     {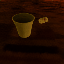} &
     \includegraphics[width=0.17\textwidth]{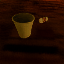} \\
     \includegraphics[width=0.17\textwidth]            {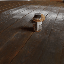} &
     \includegraphics[width=0.17\textwidth]           {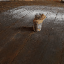} &
     \includegraphics[width=0.17\textwidth]      {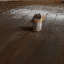} &
     \includegraphics[width=0.17\textwidth]     {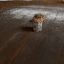} &
     \includegraphics[width=0.17\textwidth]{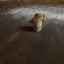} \\
  \end{tabular}
  \caption{Qualitative results on Autoencoder's reconstruction Kubric-Occlusion dataset}
  \label{fig:result-kubric}
  \vspace{-10pt}
\end{figure}

\begin{figure}
  \centering
  \small 
  \setlength{\tabcolsep}{0.5pt} 
  \begin{tabular}{c cc|cccccc} 
    & \multicolumn{2}{c}{\centering{Input}} &   \multicolumn{6}{c}{\centering{Prediction}}   \\
    & \textbf{$t=1$} & \textbf{$t=5$} & \textbf{$t=21$} & \textbf{$t=22$} & \textbf{$t=23$} & \textbf{$t=24$} & \textbf{$t=25$}\\
    \rotatebox{90}{GT} &            \includegraphics[width=0.13\textwidth]{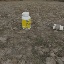} &
                                    \includegraphics[width=0.13\textwidth]{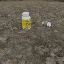} &
                                    \includegraphics[width=0.13\textwidth]{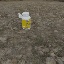} &
                                    \includegraphics[width=0.13\textwidth]{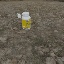} &
                                    \includegraphics[width=0.13\textwidth]{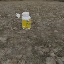} &
                                    \includegraphics[width=0.13\textwidth]{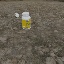} &
                                    \includegraphics[width=0.13\textwidth]{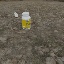} & \\
     & \multicolumn{2}{c}{SCAT}    &
                                    \includegraphics[width=0.13\textwidth]{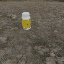} &
                                    \includegraphics[width=0.13\textwidth]{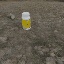} &
                                    \includegraphics[width=0.13\textwidth]{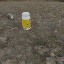} &
                                    \includegraphics[width=0.13\textwidth]{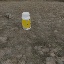} &
                                    \includegraphics[width=0.13\textwidth]{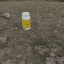} & \\
     & \multicolumn{2}{c}{SCAT-P}    &
                                    \includegraphics[width=0.13\textwidth]{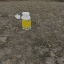} &
                                    \includegraphics[width=0.13\textwidth]{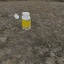} &
                                    \includegraphics[width=0.13\textwidth]{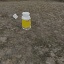} &
                                    \includegraphics[width=0.13\textwidth]{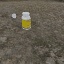} &
                                    \includegraphics[width=0.13\textwidth]{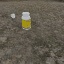} & \\
    & \multicolumn{2}{c}{SCAT-D}   &
                                    \includegraphics[width=0.13\textwidth]{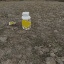} &
                                    \includegraphics[width=0.13\textwidth]{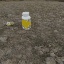} &
                                    \includegraphics[width=0.13\textwidth]{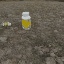} &
                                    \includegraphics[width=0.13\textwidth]{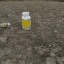} &
                                    \includegraphics[width=0.13\textwidth]{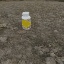} & \\
    & \multicolumn{2}{c}{SimVP}   &
                                    \includegraphics[width=0.13\textwidth]{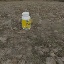} &
                                    \includegraphics[width=0.13\textwidth]{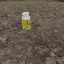} &
                                    \includegraphics[width=0.13\textwidth]{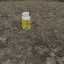} &
                                    \includegraphics[width=0.13\textwidth]{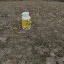} &
                                    \includegraphics[width=0.13\textwidth]{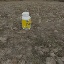} & \\
    & \multicolumn{2}{c}{SCAT-PD}   &
                                    \includegraphics[width=0.13\textwidth]{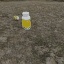} &
                                    \includegraphics[width=0.13\textwidth]{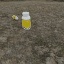} &
                                    \includegraphics[width=0.13\textwidth]{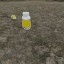} &
                                    \includegraphics[width=0.13\textwidth]{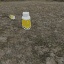} &
                                    \includegraphics[width=0.13\textwidth]{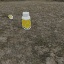} & \\
  \end{tabular}
  \caption{Video prediction example on Kubric-Occlusion (1)}
  \label{fig:result-kubric}
  \vspace{-10pt}
\end{figure}

  \begin{figure}
  \centering
  \small 
  \setlength{\tabcolsep}{0.5pt} 
  \begin{tabular}{c cc|cccccc} 
    & \multicolumn{2}{c}{\centering{Input}} &   \multicolumn{6}{c}{\centering{Prediction}}   \\
    & \textbf{$t=1$} & \textbf{$t=5$} & \textbf{$t=12$} & \textbf{$t=14$} & \textbf{$t=16$} & \textbf{$t=18$} & \textbf{$t=20$}\\
    \rotatebox{90}{GT} &            \includegraphics[width=0.13\textwidth]{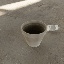} &
                                    \includegraphics[width=0.13\textwidth]{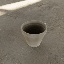} &
                                    \includegraphics[width=0.13\textwidth]{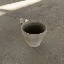} &
                                    \includegraphics[width=0.13\textwidth]{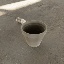} &
                                    \includegraphics[width=0.13\textwidth]{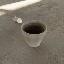} &
                                    \includegraphics[width=0.13\textwidth]{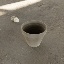} &
                                    \includegraphics[width=0.13\textwidth]{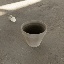} & \\
     & \multicolumn{2}{c}{SCAT}    &
                                    \includegraphics[width=0.13\textwidth]{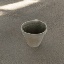} &
                                    \includegraphics[width=0.13\textwidth]{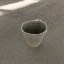} &
                                    \includegraphics[width=0.13\textwidth]{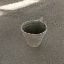} &
                                    \includegraphics[width=0.13\textwidth]{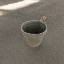} &
                                    \includegraphics[width=0.13\textwidth]{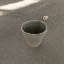} & \\
     & \multicolumn{2}{c}{SCAT-P}    &
                                    \includegraphics[width=0.13\textwidth]{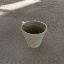} &
                                    \includegraphics[width=0.13\textwidth]{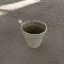} &
                                    \includegraphics[width=0.13\textwidth]{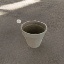} &
                                    \includegraphics[width=0.13\textwidth]{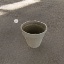} &
                                    \includegraphics[width=0.13\textwidth]{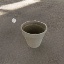} & \\
    & \multicolumn{2}{c}{SCAT-D}   &
                                    \includegraphics[width=0.13\textwidth]{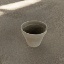} &
                                    \includegraphics[width=0.13\textwidth]{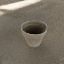} &
                                    \includegraphics[width=0.13\textwidth]{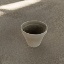} &
                                    \includegraphics[width=0.13\textwidth]{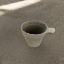} &
                                    \includegraphics[width=0.13\textwidth]{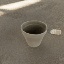} & \\
    & \multicolumn{2}{c}{SimVP}   &
                                    \includegraphics[width=0.13\textwidth]{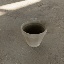} &
                                    \includegraphics[width=0.13\textwidth]{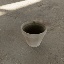} &
                                    \includegraphics[width=0.13\textwidth]{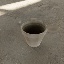} &
                                    \includegraphics[width=0.13\textwidth]{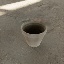} &
                                    \includegraphics[width=0.13\textwidth]{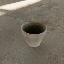} & \\
    & \multicolumn{2}{c}{SCAT-PD}   &
                                    \includegraphics[width=0.13\textwidth]{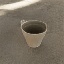} &
                                    \includegraphics[width=0.13\textwidth]{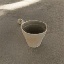} &
                                    \includegraphics[width=0.13\textwidth]{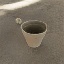} &
                                    \includegraphics[width=0.13\textwidth]{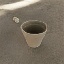} &
                                    \includegraphics[width=0.13\textwidth]{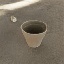} & \\
  \end{tabular}
  \caption{Video prediction example on Kubric-Occlusion (2)}
  \label{fig:result-kubric}
  \vspace{-10pt}
\end{figure}
\begin{figure}
  \centering
  \small 
  \setlength{\tabcolsep}{0.5pt} 
  \begin{tabular}{c cc|cccccc} 
    & \multicolumn{2}{c}{\centering{Input}} &   \multicolumn{6}{c}{\centering{Prediction}}   \\
    & \textbf{$t=1$} & \textbf{$t=5$} & \textbf{$t=6$} & \textbf{$t=7$} & \textbf{$t=8$} & \textbf{$t=9$} & \textbf{$t=10$}\\
    \rotatebox{90}{GT} &            \includegraphics[width=0.13\textwidth]{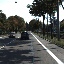} &
                                    \includegraphics[width=0.13\textwidth]{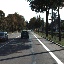} &
                                    \includegraphics[width=0.13\textwidth]{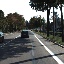} &
                                    \includegraphics[width=0.13\textwidth]{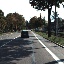} &
                                    \includegraphics[width=0.13\textwidth]{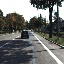} &
                                    \includegraphics[width=0.13\textwidth]{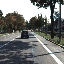} &
                                    \includegraphics[width=0.13\textwidth]{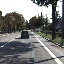} & \\
     & \multicolumn{2}{c}{SCAT}    &
                                    \includegraphics[width=0.13\textwidth]{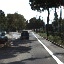} &
                                    \includegraphics[width=0.13\textwidth]{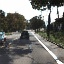} &
                                    \includegraphics[width=0.13\textwidth]{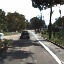} &
                                    \includegraphics[width=0.13\textwidth]{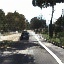} &
                                    \includegraphics[width=0.13\textwidth]{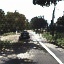} & \\
     & \multicolumn{2}{c}{SCAT-P}    &
                                    \includegraphics[width=0.13\textwidth]{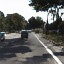} &
                                    \includegraphics[width=0.13\textwidth]{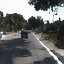} &
                                    \includegraphics[width=0.13\textwidth]{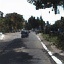} &
                                    \includegraphics[width=0.13\textwidth]{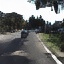} &
                                    \includegraphics[width=0.13\textwidth]{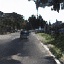} & \\
    & \multicolumn{2}{c}{SCAT-D}   &
                                    \includegraphics[width=0.13\textwidth]{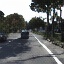} &
                                    \includegraphics[width=0.13\textwidth]{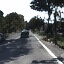} &
                                    \includegraphics[width=0.13\textwidth]{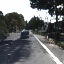} &
                                    \includegraphics[width=0.13\textwidth]{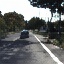} &
                                    \includegraphics[width=0.13\textwidth]{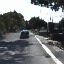} & \\
    & \multicolumn{2}{c}{SimVP}   &
                                    \includegraphics[width=0.13\textwidth]{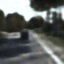} &
                                    \includegraphics[width=0.13\textwidth]{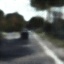} &
                                    \includegraphics[width=0.13\textwidth]{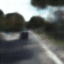} &
                                    \includegraphics[width=0.13\textwidth]{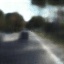} &
                                    \includegraphics[width=0.13\textwidth]{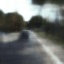} & \\
    & \multicolumn{2}{c}{SCAT-PD}   &
                                    \includegraphics[width=0.13\textwidth]{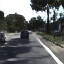} &
                                    \includegraphics[width=0.13\textwidth]{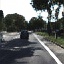} &
                                    \includegraphics[width=0.13\textwidth]{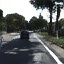} &
                                    \includegraphics[width=0.13\textwidth]{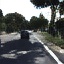} &
                                    \includegraphics[width=0.13\textwidth]{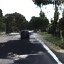} & \\
  \end{tabular}
  \caption{Video prediction example on KITTI (1)}
  \label{fig:result-kubric}
  \vspace{-10pt}
\end{figure}
\begin{figure}
  \centering
  \small 
  \setlength{\tabcolsep}{0.5pt} 
  \begin{tabular}{c cc|cccccc} 
    & \multicolumn{2}{c}{\centering{Input}} &   \multicolumn{6}{c}{\centering{Prediction}}   \\
    & \textbf{$t=1$} & \textbf{$t=5$} & \textbf{$t=6$} & \textbf{$t=7$} & \textbf{$t=8$} & \textbf{$t=9$} & \textbf{$t=10$}\\
    \rotatebox{90}{GT} &            \includegraphics[width=0.13\textwidth]{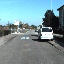} &
                                    \includegraphics[width=0.13\textwidth]{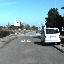} &
                                    \includegraphics[width=0.13\textwidth]{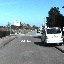} &
                                    \includegraphics[width=0.13\textwidth]{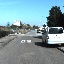} &
                                    \includegraphics[width=0.13\textwidth]{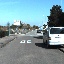} &
                                    \includegraphics[width=0.13\textwidth]{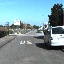} &
                                    \includegraphics[width=0.13\textwidth]{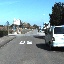} & \\
     & \multicolumn{2}{c}{SCAT}    &
                                    \includegraphics[width=0.13\textwidth]{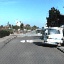} &
                                    \includegraphics[width=0.13\textwidth]{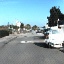} &
                                    \includegraphics[width=0.13\textwidth]{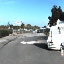} &
                                    \includegraphics[width=0.13\textwidth]{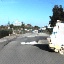} &
                                    \includegraphics[width=0.13\textwidth]{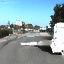} & \\
     & \multicolumn{2}{c}{SCAT-P}    &
                                    \includegraphics[width=0.13\textwidth]{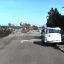} &
                                    \includegraphics[width=0.13\textwidth]{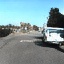} &
                                    \includegraphics[width=0.13\textwidth]{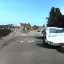} &
                                    \includegraphics[width=0.13\textwidth]{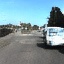} &
                                    \includegraphics[width=0.13\textwidth]{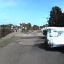} & \\
    & \multicolumn{2}{c}{SCAT-D}   &
                                    \includegraphics[width=0.13\textwidth]{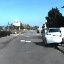} &
                                    \includegraphics[width=0.13\textwidth]{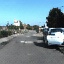} &
                                    \includegraphics[width=0.13\textwidth]{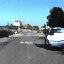} &
                                    \includegraphics[width=0.13\textwidth]{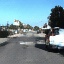} &
                                    \includegraphics[width=0.13\textwidth]{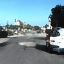} & \\
    & \multicolumn{2}{c}{SimVP}   &
                                    \includegraphics[width=0.13\textwidth]{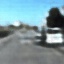} &
                                    \includegraphics[width=0.13\textwidth]{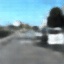} &
                                    \includegraphics[width=0.13\textwidth]{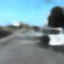} &
                                    \includegraphics[width=0.13\textwidth]{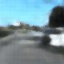} &
                                    \includegraphics[width=0.13\textwidth]{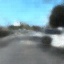} & \\
    & \multicolumn{2}{c}{SCAT-PD}   &
                                    \includegraphics[width=0.13\textwidth]{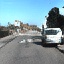} &
                                    \includegraphics[width=0.13\textwidth]{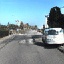} &
                                    \includegraphics[width=0.13\textwidth]{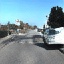} &
                                    \includegraphics[width=0.13\textwidth]{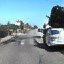} &
                                    \includegraphics[width=0.13\textwidth]{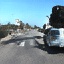} & \\
  \end{tabular}
  \caption{Video prediction example on KITTI (2)}
  \label{fig:result-kubric}
  \vspace{-10pt}
\end{figure}

\end{document}